\documentclass[10pt,twocolumn,letterpaper]{article}

\usepackage{graphicx}
\usepackage{amsmath}
\usepackage{amssymb}
\usepackage{booktabs}
\usepackage{bm}

\usepackage{array}

\usepackage[pagebackref,breaklinks,colorlinks]{hyperref}

\usepackage[capitalize]{cleveref}
\crefname{section}{Sec.}{Secs.}
\Crefname{section}{Section}{Sections}
\Crefname{table}{Table}{Tables}
\crefname{table}{Tab.}{Tabs.}

\begin{document}

\title{Learning AND-OR Templates for Professional Photograph Parsing and Guidance}

\author{
Xin Jin, Liaoruxing Zhang, Chenyu Fan, Wenbo Yuan\\
Beijing Electronic Science and Technology Institute, Beijing, China\\
{\tt\small jinxinbesti@foxmail.com, zhangxingxing002@gmail.com, 3497961491@qq.com, weber\_yuan@163.com}
}

\maketitle

\begin{abstract}
   Since the development of photography art, many so-called ”templates” have been formed, namely visual styles summarized from a series of themed and stylized photography works. In this paper, we propose to analysize and and summarize these ’templates’ in photography by learning composite templates of photography images. We present a framework for learning a hierarchical reconfigurable image template from photography images to learn and characterize the ”templates” used in these photography images. Using this method, we measured the artistic quality of photography on the photos and conducted photography guidance. In addition, we also utilized the ”templates” for guidance in several image generation tasks. Experimental results show that the learned templates can well describe the photography techniques and styles, whereas the proposed approach can assess the quality of photography images as human being does.
\end{abstract}

\section{Introduction}

Since the development of photography, many "templates" have emerged. These so-called templates are visual styles summarized from a series of themed and stylized photography works \cite{2013Digital,2015Photography}. These templates typically comprise elements such as color, composition, shooting angle, and subjects. By combining and adjusting these elements, a unique visual effect and style can be created, enabling photographers to better express their creativity and artistic vision in the shooting process \cite{0PHOTOGRAPHY,2014The,2015Introduction}.

\begin{figure}[htbp]
	\centering
	\includegraphics[width=\linewidth]{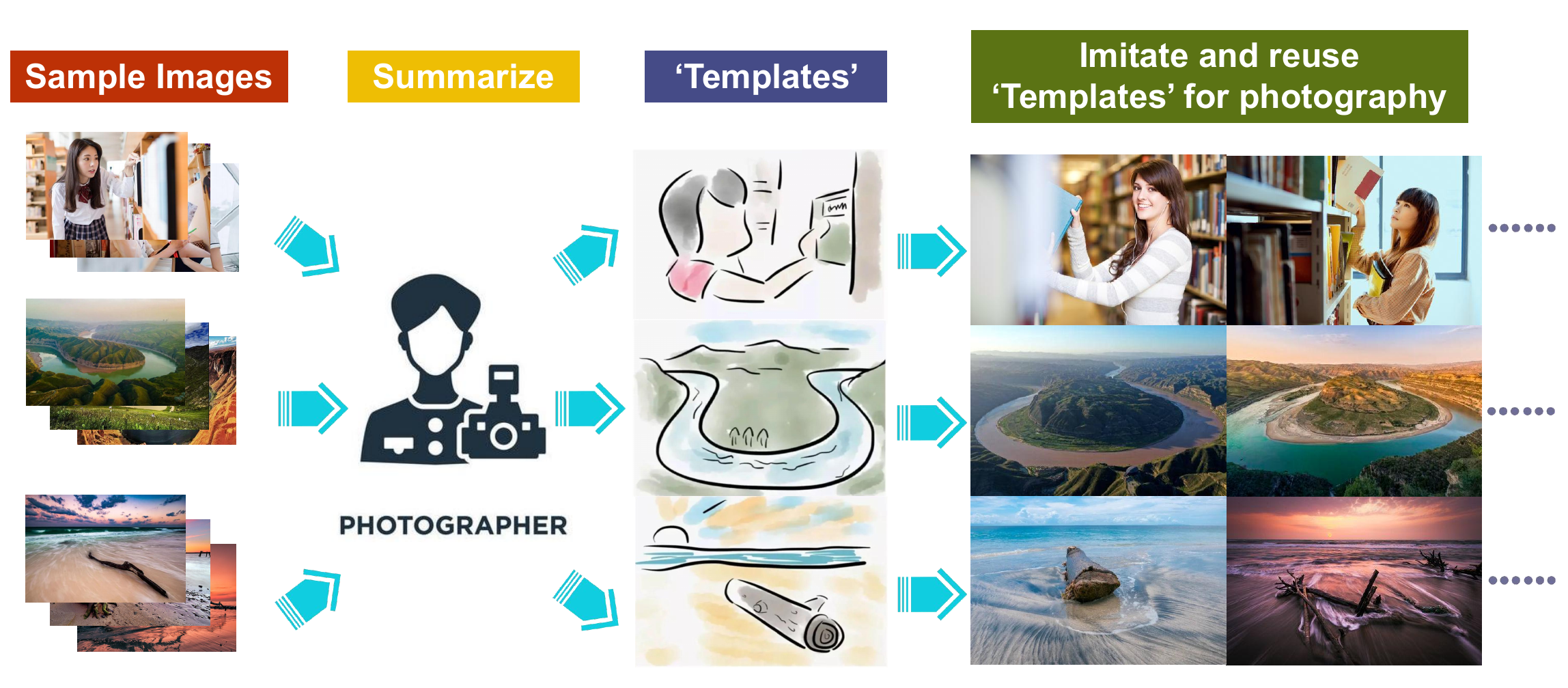}
	\caption{The process of summarizing the 'template' by photographers. Photographers can summarize templates from a series of themed and stylized photography works, and then use these templates as a reference to improve and innovate, forming their own unique shooting style and aesthetic.}
	\label{introduct}
\end{figure}

Moreover, these templates can be referenced and utilized by other photographers or photography enthusiasts to aid them in realizing their shooting goals and creativity \cite{1999Using,1986Visual}. They can also be further improved and innovated upon to develop their own distinctive visual style and aesthetic \cite{2004Representing}. As is shown in Fig. \ref{introduct}, photographers can extract templates from a series of themed and stylish photography works, and then emulate and apply them to their own photography.

However, these templates are static knowledge that photographers summarize based on their own experience. How to automatically discover the photography paradigms used from massive internet images is a very important new task. The goal of our paper is to learn these "templates" in photography and present them as a template image. For photography, the automatically discovered photography mode can verify the photography rules summarized by human artists, and can dynamically discover new photography paradigms.

To achieve the task of automatically discovering photography paradigms in images, we propose a framework for learning a reconfigurable composite template. By training a small batch of photography images with the same theme, our template can learn the paradigm used for this type of photography image. Our template consists of multiple AND nodes and OR nodes. The AND nodes represent compositions of parts, while the OR nodes account for articulation and structural variation of parts. We reflect the configuration of the captured objects in the image by combining AND nodes with OR nodes.

Besides, photographiy works usually contain multiple shooting objects, and there are multiple complex relationships between the shooting objects. To address this challenge, our template adopts a two-layer inference structure: i) learning templates of objects that make up photography images. ii) learning scene templates based on the corresponding object templates. We will introduce the specific template learning methods in Session 4.

The templates are further used to photography analysis. We utilize templates for image interpretability evaluation and guide photographic images using our templates. Finally, we explore the application of templates in image design such as movie poster design.

The main contributions of our paper can be summarized as follow:

\begin{itemize}
	\item We propose a framework to learn art templates for photography through a set of photographic images.
	\item We utilize templates for image interpretability evaluation and guide photographic images using our templates.
	\item We explore the application of templates in image design such as movie poster design. 
\end{itemize}

\section{Related Work}
In this paper, we focus on photography analysis including composite template learning, quantitative assessment and explainable photography guidance. Roughly, the methods related to our topic can be divided into two aspects: image template, photography assessment and guidance.

\subsection{Image templates}

Image templates, especially deformable ones, have been extensively studied for detection, recognition and tracking, for example, deformable templates \cite{1992Feature}, active appearance models \cite{2004Automatic}, pictorial structures \cite{2005Pictorial}, constellation model \cite{2007Weakly}, part-based latent SVM model \cite{Felzenszwalb2010Object}, recursive compositional model \cite{2010Part}, region-based taxonomy \cite{2007}, hierarchical parts dictionary \cite{2007Towards} and Active Basis model \cite{Ying2010Learning}. 

Zhu et al. proposed HIT model \cite{2012Learning} for object detection tasks. This template is closely related to the Active Basis model which only uses sketches to represent object shapes. In contrast, the HIT model integrates texture, flatness and color features and is more expressive. Besides, HIT model is also related to the primal sketch representation which combines sketchable (primitives) and non-
sketchable (textures and flatness) image components. The difference is that the primal sketch is a low-middle level visual representation for generic images while the HIT model is a high level vision representation for objects that have similar configurations in a category.

In order to cope with more complex images, Si and Zhu et al. proposed the AND-OR template \cite{Zhangzhang2013Learning} based on the HIT model. An AND-OR template, including AND nodes and OR nodes, is a stochastic reconfigurable template that generates a set of valid configurations or object templates. The AND nodes represent compositions of parts, while the OR nodes account for articulation and structural variation of parts. The AND-OR template can cope with more complex detection tasks of image content, effectively compensating for the shortcomings of the HIT model.

In the field of image aesthetics evaluation, Diep et al. construct an image composition template dataset(PoB) \cite{2018PoB}. The PoB dataset consists of 10000 paintings and 4959 photos, with each image labeled with a composition template. This dataset lays the foundation for the discovery of aesthetic patterns and the expression of aesthetic knowledge.

Our composite template is closely related to the AND-OR model which generates a set of valid configurations or object templates. Compared to the PoB dataset, our template can not only reflect the composition of the image, but also learn about the multiple attributes of the image such as color and texture.

\subsection{Photography Assessment and Guidance}
Photo-quality assessment are popular and extremely challenging topics. Many interesting applications including perceptual photo-quality assessment, photo view recommendation and quality-based photo re-ranking have drawn great attention in the multimedia research community.

Luo el al \cite{2008Photo}. collected 17613 photos with manually labelled ground truth. They divided the photos into seven categories based on the photo contents, and developed a set of subject area extraction methods and visual features, which are specially designed for different categories. This method greatly improves photo quality assessment performance.

Li et al \cite{2010Aesthetic}. proposed a framework of automatically evaluating the aesthetic quality on the photos with faces. Jin et al \cite{2021Aesthetic}. propose the aesthetic dashboard which shows the rich aesthetic evaluation and guidance for the mobile users to shoot excellent photos. Users can refer the template matching scores from the aesthetic dashboard to obtain the desired patterns of light, color and composition.

While these methods have made significant strides in evaluating and guiding photography in various aspects, a comprehensive method for photography assessment and guidance is yet to be established. In addition, these methods rely on a large amount of training data, and the results are inexplicable, which cannot meet the needs of technological development. 

\section{Our Method}
In this section, we will specifically introduce our method for learning composite templates from professional photography images, mainly including the representation of composite templates and template learning algorithm. 

\subsection{Representation of Our Composite Templates}

Due to the complexity of scene images, our template representation includes both object templates and the scene templates. We first represent templates for each object that makes up the scene, and then combine these object templates into corresponding scene templates.

The representation of object templates adopts an AND-OR graph structure. The AND nodes represent compositions of parts, while the OR nodes account for articulation and structural variation of parts. A composite template consists of a number of configurations of parts, which include:

i) structural variabilities (i.e. what parts appear);

ii) geometric variabilities (i.e. where they appear).

\begin{figure*}[t]
	\centering
	\includegraphics[width=\linewidth]{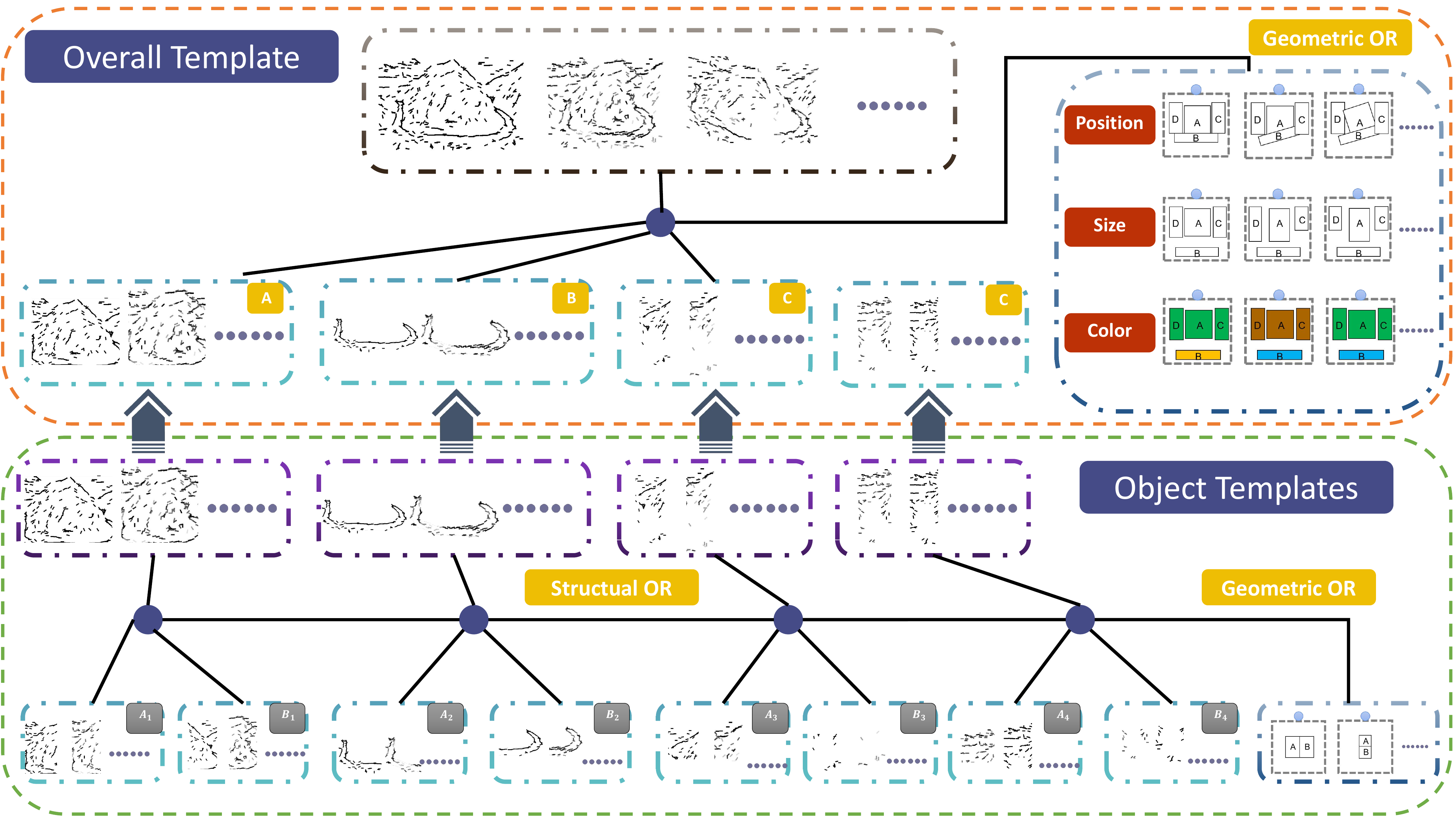}
	\caption{The detailed illustration of our composite template. For OR nodes, we illustrate typical templates with structural variations and geometric transforms. The AND nodes are denoted by solid blue circles. The terminal nodes are individual parts which are represented by professional photography templates.}
	\label{structure}
\end{figure*}

Taking a scene composed of a river around the mountain as an example, this scene mainly includes four objects: river, middle mountain, left mountain and right mountain. We need to represent the templates for these two objects separately. Fig. \ref{structure} illustrates the scene templates for river and mountains. The terminal nodes are shown as light blue rectangles. AND nodes are denoted as blue solid circles. OR nodes (for both geometric and structural variabilities) are drawn with dashed boxes together with typical configurations. The root node is an OR node with all the valid configurations of the scene. It is branched into several sets of valid structural configurations as well as geometric configurations (represented as AND nodes) of four sub-parts: middle mountain(A), river(B), left mountain(C) and right mountain(D). As we move down the composite template, middle mountain and river can be separated into left and right parts. Similarly, left mountain and right mountain can also be divided into up and down parts. These parts are displayed as terminal nodes in the object template. The object template represents these parts as terminal nodes, whereas the scene template represents the object templates that constitute the scene as terminal nodes. The reason we do not consider these parts as terminal nodes in our scene template is because they make up a relatively small portion of the overall scene. Starting from such small parts would lead to a large number of unnecessary configurations, which would adversely impact the efficiency of our template learning process.

We use a stochastic context free grammar to regulate the structural and geometric variabilities in object templates. It can efficiently capture high-order interaction of parts and compositionality.

In the representation of terminal nodes, we propose the PATs (photography art templates) based on method in . 
A PAT is specified by a list:

\begin{center}
	PAT = \{($B_1$, $x_1$, $y_1$, $s_1$, $o_1$), ($h_2$, $x_2$, $y_2$), ($h_3$, $x_3$,$y_3$), ($B_4$, $x_4$, $y_4$, $s_4$, $o_4$), ...\}
\end{center}
where $B_1$, $B_4$, ... are image primitives and $h_2$, $h_3$, ... are histogram descriptors for texture, flatness and color. {($x_j$, $y_j$)} denote the selected locations and {$o_j$} are the selected orientations. 

In addition, the structural and geometric configurations are not observed in training images, and thus are modeled by two separate sets of latent random variables:

i) \textbf{Structural configuration}. We set s to represent structural configurations in object or scene templates. In object templates, the structual configuration s is a binary activation vector of length K (Number of Terminal Nodes), indicating which parts are activated. $s_k$ = 1 means node is activated and appears in the object image. In scene templates, the structual configuration s is a binary activation vector of length num (num = |$\Delta ^{(3)}$|), indicating which objects are activated. $s'_{num}$ = 1 means object is activated and appears in the image.

ii) \textbf{Geometric configuration}. We set g to represent geometric configurations in object or scene templates. In object templates, the geometric configuration g is a list of transforms (translation, rotation and scaling) applied to the parts in terminal nodes. In scene templates, the geometric configuration g is a list containing basic attributes applied to objects (position, size, and color) and image composition information.

We use a stochastic context free grammar for template representation and construction. In stochastic context free grammar, each possible paradigm is assigned a probability. The probability model used in our template is represented as follows:

Let $\chi_+$ = {$I_1$,...,$I_N$} be positive example images (e.g. objects in object templates) governed by the underlying target distribution f(I). Let $\chi_-$ be a large set of generic natural images governed by the reference distribution q(I). Our objective of learning is to pursue a model p(I) to approximate f(I) in a series of steps:

\begin{equation}
	q(I) = p_0(I) \rightarrow p_1(I) \rightarrow ···p_T(I) = p(I) \thickapprox f(I)
\end{equation}
starting from q.

The model p after T iterations contains T selected features {$r_t$ : t = 1, ..., T }. If the selected feature responses capture all information about image I, it can be shown by variable transformation that:

\begin{equation}
	\frac{p(I)}{q(I)} = \frac{p(r_1,...,r_T)}{p(r_1,...,r_T)}
\end{equation}
So p can be constructed by reweighting q with the marginal likelihood ratio on selected features.

Under the maximum entropy principle, p(I) can be expressed in the following log-linear form:

\begin{equation}
	p(I) = q(I) \prod_{t = 1}^{T}[\frac{1}{z_t} exp\{\beta_t r_t(I)\}]
\end{equation}
where $\beta_t$ is the parameter for the t-th selected feature $r_t$ and $z_t$ ($z_t$ $\geq$ 0) is the individual normalization constant
determined by $\beta_t$:

\begin{equation}
	z_t = \sum_{r_t}q(r_t)exp\{\beta_t r_t \}
\end{equation}

Based on this probability model, each configuration (s, g) in the template has a corresponding probability.

The complete likelihood for the template is defined as
\begin{equation}
	p(I,s,g|Temp,\beta) = p(s,g|Temp) \cdot p(I|s,g,\beta)
\end{equation}
and the image likelihood conditioned on the configuration (s,g) is a log-linear form following Eq. (6):
\begin{equation}
	p(I|s,g,\beta) = exp\{\sum_{k = 1}^{K}s_k(\sum_{j = 1}^{D}\beta_{k,j}r_j(I) - \log Z_k)\}q(I)
\end{equation}
Where D denotes the total number of feature responses, $\beta$ is the parameter for the t-th selected feature $r_t$ and $z_t$ ($z_t$ $\geq$ 0) is the individual normalization constant determined by $\beta$. q(I) represents the reference distribution. 

We shall call the log-likelihood ratio $\log \frac{p}{q}$ as a template matching score, which measures the information gain of using this PPT to represent the object:
\begin{equation}
	Score(I) = \log \frac{p(I|s,g,\beta)}{q(I)} = \sum_{k = 1}^{K}Score(PAT_k,I)
\end{equation}
where
\begin{equation}
	Score(PAT_k,I) = s_k(\sum_{j = 1}^{D}\beta_{k,j}r_j(I) - \log Z_k)
\end{equation}

\subsection{Learning Composite Template From Photography Images}

\begin{figure*}[htbp]
	\centering
	\includegraphics[width=\linewidth]{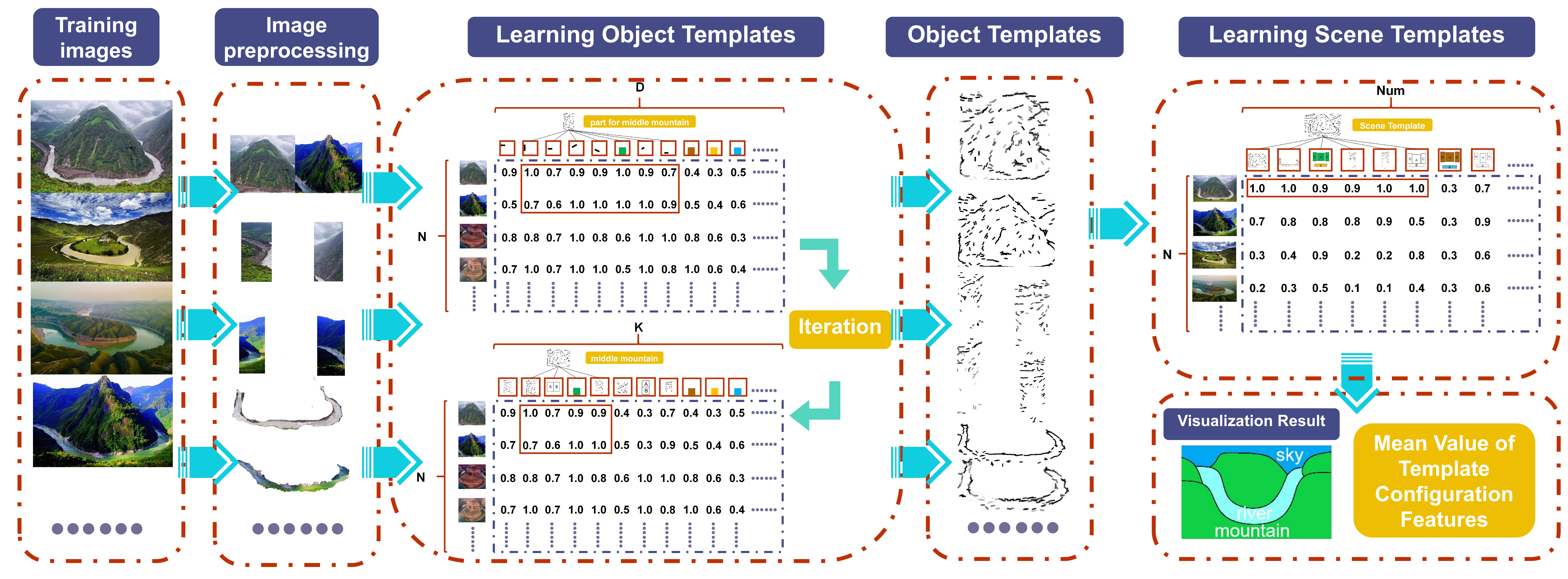}
	\caption{Overall process of template learning.}
	\label{flow}
\end{figure*} 

The learning process of our composite template includes two aspects: 

i). learning each object template. 

ii). combining the learned object template to learn the scene template.

We choose a set of photographic images from the same theme as training data. We first preprocess the training images using the YOLOv8 model to detect and segment the objects that make up each image. Then, we use an EM-type block pursuit algorithm to learn the terminal nodes and the non-terminal nodes in object templates from training images.

The learning of object templates is a block pursuing process (i.e. activations of parts in objects). So we use an EM-type block pursuit algorithm. Because our template learning is divided into two parts: object template learning and scene template learning, the learning is performed on the data matrix R shown in Fig. \ref{matrix1}. 

Each row of R is a feature vector for an object image in training images. Therefore R is a matrix with N (number of positive examples) rows and D (number of all candidate features) columns, and each entry $R_{ij}$ = $r_j$($I_i$) is a feature response (0 $\leq$ $R_{ij}$ $\leq$ 1). Larger value of $R_{ij}$ means feature j appears in image $I_i$ with higher probability.

On the data matrix, we pursue large blocks \{ $B_k$ : k = 1, ..., K \} with lots of 1’s, which correspond to PPTs that appear frequently and with high confidence. A block is specified by a set of common features (columns) shared by a set of examples (rows). The significance of block $B_k$ is measured by the summation over the block:
\begin{figure}[htbp]
	\centering
	\includegraphics[width=1\linewidth]{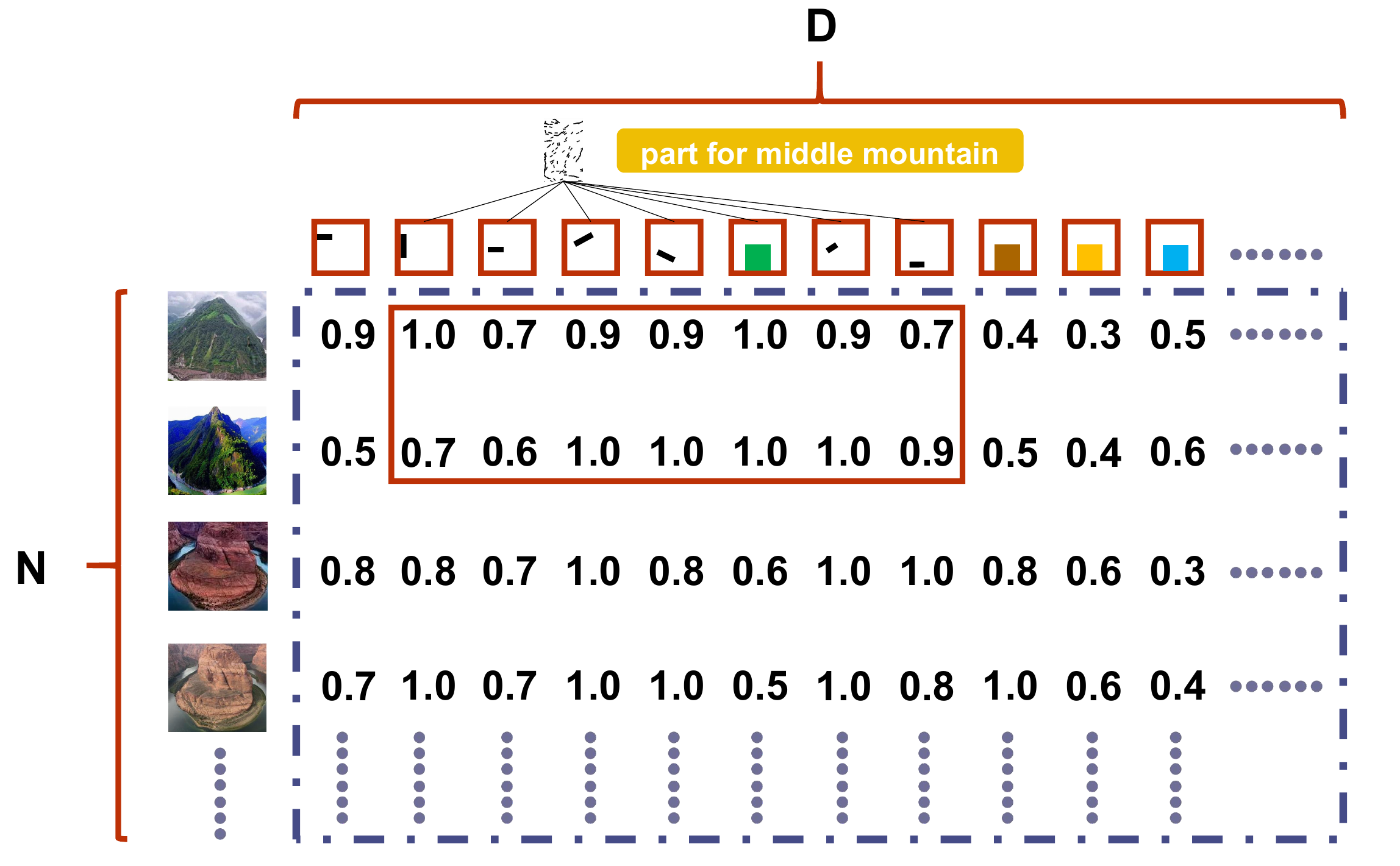}
	\caption{Initial Data Matrix. Data matrix R is a matrix with N (number of positive examples) rows and D (number of all candidate features) columns, and each entry $R_{ij}$ = $r_j$($I_i$) is a feature response (0 $\leq$ $R_{ij}$ $\leq$ 1). Larger value of $R_{ij}$ means feature j appears in image $I_i$ with higher probability.}
	\label{matrix1}
\end{figure}

\begin{equation}
	Score(B_k) = \sum_{\substack{i \in rows(B_k) \\ j \in cols(B_k)}}(\beta_{k,j}R_{i,j} - \log z_{k,j})
\end{equation}
where rows and cols denote the rows and columns of block $B_k$. cols($B_k$) corresponds to the selected features in $PAT_k$; and rows
($B_k$) are the examples on which $PAT_k$ is activated. $\beta_k$, j is the multiplicative parameter of feature j in $PAT_k$, and $z_k$, j is the individual normalizing constant determined by $\beta_k$, j. See Eq.4 for estimation of $z_k$, j.

The score of $B_k$ is equal with the summation of Eq.4 in the main manuscript over positive examples \{$I_i$ : i = 1, ..., N\}: 

The process of block tracking is not endless. Pursuing blocks by maximizing the total score need correspond to maximum likelihood estimation and the information projection principle. So,the block pursuit is a penalized maximum likelihood estimation problem, minimizing a two-term cost function:

\begin{center}
	$-L(R,\beta,s,g) + penalty(\beta)$
\end{center}

We also enforce that the coefficients $\beta_{k,:}$ of $PAT_k$ is confined within a local region of the object window, and for each local region exactly one block is activated for each image example, so that the activated PPTs do not overlap. The learned blocks can be ranked by the score (or information gain) in Eq.13 and the blocks with small scores are discarded. After T iterations, we can obtain the object template's Coefficient matrix $\beta^{(T)}$.

The same algorithm is applied recursively to learn the overall template. Similar to object template learning, the overall template learning is also performed on the data matrix R'. Each row of R' is a feature vector for a photography image in $\chi_+$. Therefore R' is a matrix with N' (number of positive examples) rows and D' (number of all candidate objects for photography images) columns, and each entry $R'_{ij}$ = $r'_j$($I'_i$) is a feature response (0 $\leq$ $R'_{ij}$ $\leq$ 1). Larger value of $R'_{ij}$ means object j appears in image $I'_i$ with higher probability.

On the data matrix for overall template, we also pursue large blocks \{ $B'_{num}$ : num = 1, ..., Num \} with lots of 1’s, which correspond to object templates that appear frequently and with high confidence. A block is specified by a set of objects (columns) shared by a set of examples (rows). So we can reuse the EM-type block pursuit algorithm used in object template learning into the scene template learning.

Finally, we take the mean of all learned possible template configuration features and output them as visualization results. We show the overall template learning process in the Fig. \ref{flow}.

\begin{figure*}[htbp]
	\centering
	\includegraphics[width=1\linewidth]{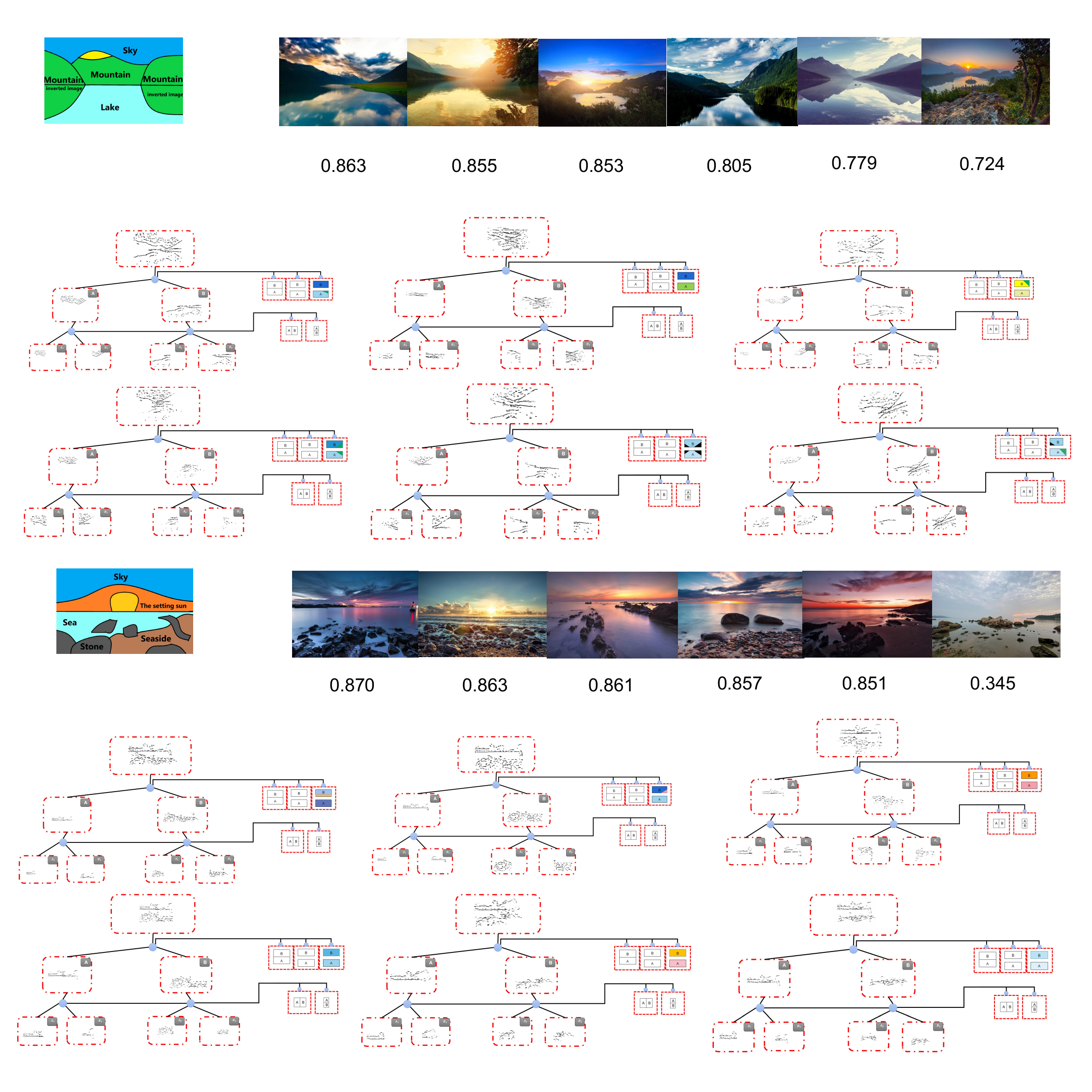}
	\caption{Results of quantitative assessment. In addition to being able to output the matching score between the image and the template, we can also output the configuration of the image on the template.}
	\label{result2}
\end{figure*}
\begin{table*}[h]
	\centering
	\caption{Quantitative comparison with the methods using single input image on the DeepFashion dataset }
	\label{tab:comp}
	\resizebox{!}{21mm}{
		\begin{tabular}{ccccc}
			\toprule
			{\footnotesize Method} & {\footnotesize Accuracy} & {\footnotesize Number of Parameters} & {\footnotesize Training Complexity} & {\footnotesize Srcc}\\
			\midrule
			{\footnotesize Our Method} & {\footnotesize $\bm{85.65 \%}$} & {\footnotesize $\bm{2.3 \times 10^3}$} & {\footnotesize $\bm{6.7 \times 10^4}$} & {\footnotesize $\bm{0.8419}$}\\
			{\footnotesize AOT } & {\footnotesize 83.09 \%} & {\footnotesize $1.9 \times 10^3$} & {\footnotesize $5.6 \times 10^4$} & {\footnotesize 0.8274} \\
			{\footnotesize VGG19 } & {\footnotesize 78.08 \%} & {\footnotesize $1.4 \times 10^8$} & {\footnotesize $4.8 \times 10^{14}$} & {\footnotesize 0.7506}\\
			{\footnotesize ResNet50 } & {\footnotesize 85.71 \%} & {\footnotesize $2.5 \times 10^8$} & {\footnotesize $7.8 \times 10^{14}$} & {\footnotesize 0.7842} \\
			{\footnotesize VGG19 + templates} & {\footnotesize 81.43 \%} & {\footnotesize $1.7 \times 10^8$} & {\footnotesize $5.1 \times 10^{14}$} & {\footnotesize 0.7893}\\
			{\footnotesize ResNet50 + templates} & {\footnotesize 85.87 \%} & {\footnotesize $2.9 \times 10^8$} & {\footnotesize $8.1 \times 10^{14}$} & {\footnotesize 0.8156}\\
			\bottomrule
	   \end{tabular}}
\end{table*}
\section{Experiments}
With the learned composite templates, we conduct some experiments in this session, including quantitative assessment for photography images, photography guidance and guidance for movie poster generation. In this session, we use the matching score for quantitative assessment obtained through Eq.7. This score is used for measuring the information gain of using this composite template to interpret the photographic image(e.g. the degree of matching between photography images and the learned composite template). The value of the matching score is between 0 and 1, and the closer it is to 1, the higher the degree of matching between the image and the template.

\subsection{Image Quantitative Assessment}
The matching score in our composite template learning represents the degree of matching between photographic images and this type of topic template. The matching score can be used to judge whether the photographic image conforms to this template to classify the shooting scene, and more importantly, it can be used for professional evaluation of photographic images. The templates learned from professional photography images summarize the visual style of professional photography images. The closer the photography images are to the templates of professional photography images, the higher the professional level of the images.

Therefore, we first select a set of images with high ratings on the same theme from datasets such as AVA and AADB (10-15 images). Using our method, we learn a professional template AND/OR graph for this theme from this set of images. For input images, we also learn their templates' AND/OR graphs and compare them with the professional template's AND/OR graph for the corresponding theme. We calculate the matching score that reflects the approximate degree of adherence to the professional paradigm for the given theme, which is used to evaluate the quality of the image. In addition to using matching scores for evaluation, we can also output the configuration parsing tree for each image, as shown in the Fig. \ref{result2}.

To further test the application of our template-based approach in image quality assessment, we attempted to perform an aesthetic classification task using templates on the AVA dataset \cite{2012AVA}. The AVA dataset has pre-classified images based on their aesthetic quality. Therefore, we separately learned aesthetic templates for low-quality and high-quality images. The classification of the aesthetic quality of images was accomplished by determining which template the image belonged to.

We selected 140 high-quality and 140 low-quality images from each of the 14 thematic categories in the AVA dataset as training data for our template training. For comparison, we employed commonly used deep learning classification models, VGG19 \cite{2014Very} and ResNet50 \cite{2016Deep}. In addition, we also include the commonly used image resolution method AND-OR template(AOT) \cite{Zhangzhang2013Learning} in the comparison to further demonstrate the advantages of our approach in image resolution. We split the AVA dataset into training and testing sets with an 8:2 ratio, and trained both deep learning networks separately, alongside our template method, for the classification task. Furthermore, we attempted to enhance the classification performance by incorporating the configured features extracted from our template into the deep learning networks.

The experimental results are shown in the table \ref{tab:comp}, we use FLOPs to measure the training complexity of each algorithm. It can be seen that our template method performs comparably to deep learning networks with smaller parameters and training data. Moreover, the deep learning networks incorporating the features extracted by our template outperform the baseline networks. We also calculated the correlation coefficients between the quality scores obtained from the template and deep learning methods and the annotated scores. It is evident that the similarity scores derived from our template exhibit higher correlation with the annotated scores, aligning with common aesthetic standards. This demonstrates that our template learning approach has indeed captured some paradigms present in photographic images and can be utilized for image quality assessment. Importantly, using templates allows us to provide explanations for our ratings, which opens up possibilities for further tasks.

\subsection{Photography Guidance}

Based on the configuration parsing tree of different images, our method can point out the problems in image shooting by comparing the learned template configurations, such as lack of scenery, composition problems caused by poor size and direction of scenery, etc. Then we provide guidance on text and images based on the template configuration. Therefore, our template method can also provide guidance for photographic images.

\begin{figure*}[htbp]
	\centering
	\includegraphics[width=\linewidth]{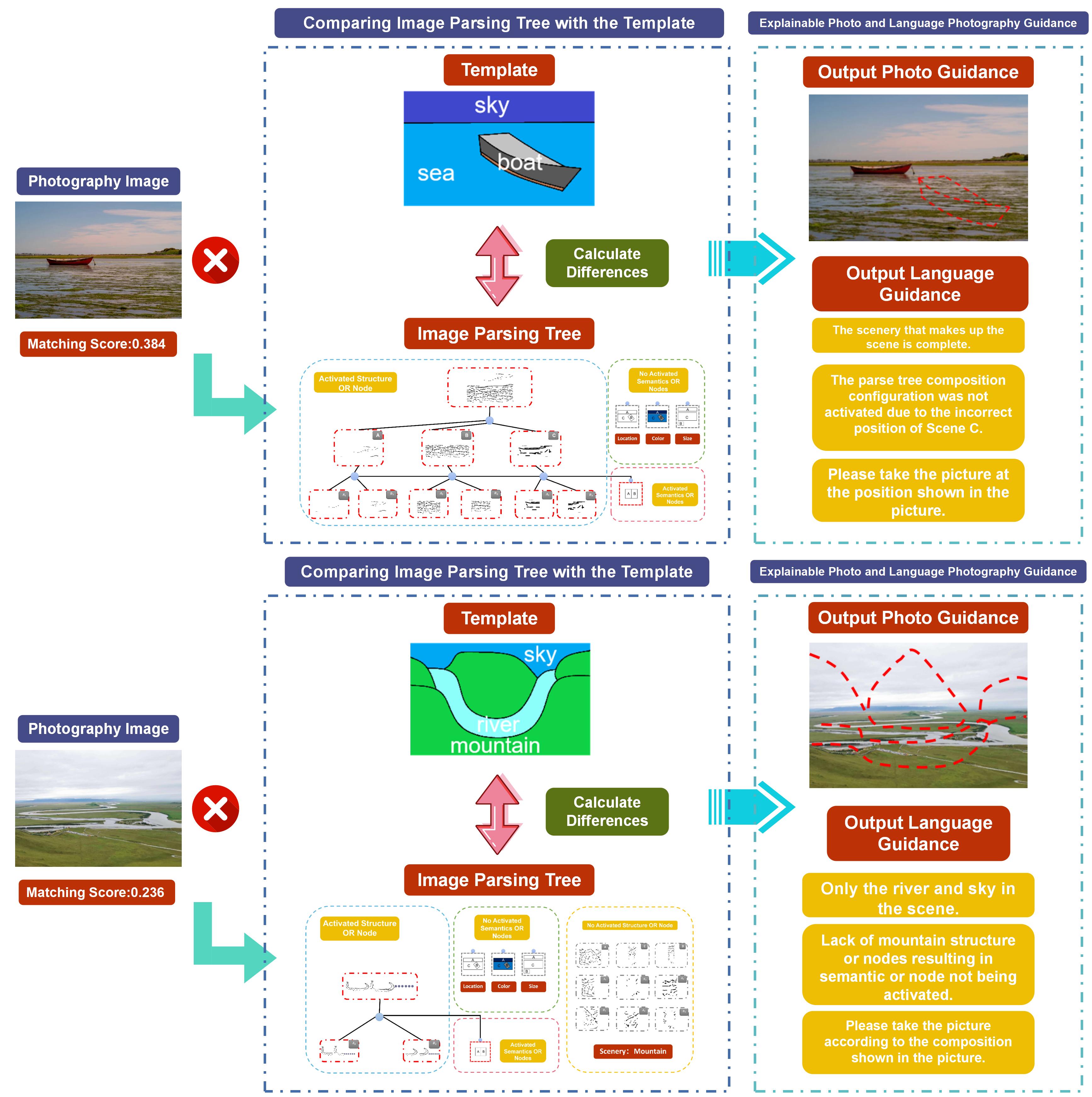}
	\caption{Interpretable photography guidance. Our method can point out the problems in photography images by comparing the learned template configurations and provide guidance on text and images based on the template configuration.}
	\label{interpret}
\end{figure*}

As is shown in Fig. \ref{interpret}, we have designed a corresponding rule corpus, which will output corresponding guiding rules based on the comparison results. At the same time, we will output a specific configuration parsing tree to indicate which part is activated for explanation. Finally, image guidance will be provided based on the visualization results of the template.




\section{Conclusion}
In this paper, we propose a framework learning art templates of photography images. We aim to automatically summarize photography templates from a series of photography works on the same theme, providing analysis and guidance for photography. Our experiments have shown that the learned templates are reasonable and assist in enhancing photographers' creativity. In future work, we plan to incorporate additional photography elements into our template, such as lighting and depth of field, to enrich the image semantics of our composite template and provide richer styles and guidance.

{\small
\bibliographystyle{ieee_fullname}
\bibliography{sample-base}
}

\end{document}